\documentclass[11pt]{article}
\usepackage[citecolor=blue]{hyperref}
\usepackage{amsmath,amsfonts,amssymb,amsthm}
\usepackage{mathrsfs}
\usepackage{verbatim}           
\usepackage{enumerate}
\usepackage[pdftex]{graphicx}
\usepackage[pdftex]{color}         
\usepackage{amscd}
\usepackage[footnotesize]{caption}

\usepackage{epstopdf}

\numberwithin{equation}{section}

\def\Z{{\mathbb Z}}
\def\R{{\mathbb R}}

\def\E{{\mathbb{E}}}            

\def\Ct{ {\mathcal{C}} }

\def\Norm{ {\mathcal{N}} }

\def\|{{|\!|}}

\newcommand{\diag}{\operatornamewithlimits{diag}}

\title{Quantifying Uncertainty in Stochastic Models with Parametric Variability.}
\author{ Kyle S. Hickmann$^{\thanks{
         Applied Mathematics and Plasma Physics, 
         Los Alamos National Laboratory,
         Los Alamos, NM, 87544; 
         hickmank@lanl.gov}}$,
     James M. Hyman$^{\thanks{
         Department of Mathematics, 
         Tulane University,
         New Orleans, LA, 70118; 
         mhyman@tulane.edu}}$,
     Sara Y. Del Valle$^{\thanks{
         Energy and Infrastructure Analysis, 
         Los Alamos National Laboratory,
         Los Alamos, NM, 87544; 
         sdelvall@lanl.gov}}$
}

\date{}
\begin{document}
\maketitle

\begin{abstract}
  We present a method to quantify uncertainty in the predictions made by
  simulations of mathematical models that can be applied to a broad class of
  stochastic, discrete, and differential equation models.  Quantifying
  uncertainty is crucial for determining how accurate the model predictions
  are and identifying which input parameters affect the outputs of interest.
  Most of the existing methods for uncertainty quantification require many
  samples to generate accurate results, are unable to differentiate where the
  uncertainty is coming from (e.g., parameters or model assumptions), or
  require a lot of computational resources. Our approach addresses these
  challenges and opportunities by allowing different types of uncertainty,
  that is, uncertainty in input parameters as well as uncertainty created
  through stochastic model components. This is done by combining the
  Karhunen-Loeve decomposition, polynomial chaos expansion, and Bayesian
  Gaussian process regression to create a statistical surrogate for the
  stochastic model. The surrogate separates the analysis of variation arising
  through stochastic simulation and variation arising through uncertainty in
  the model parameterization.  We illustrate our approach by quantifying the
  uncertainty in a stochastic ordinary differential equation epidemic model.
  Specifically, we estimate four quantities of interest for the epidemic model
  and show agreement between the surrogate and the actual model results.
\end{abstract}

\medskip


\medskip


\section{Introduction}
 
The uncertainty created by the stochastic processes and approximate parameters
in mathematical models must be quantified to assess the reliability of the
model predictions. As the complexity of models increases to include more
detail, so does the number of parameters that must be estimated. A
sophisticated framework is required to quantifying the uncertainty created by
nonlinear interactions between parameters and stochastic forces from a limited
number of computational experiments. We will describe an approach for
uncertainty quantification (UQ) for the predicted output of a simulation,
referred to as \emph{quantities of interest} (QOI).

If computer time is not an issue, then information about the predicted
distribution of QOI can be extracted by a traditional Monte Carlo approach
\cite{robert1999monte}. In traditional Monte Carlo, a comprehensive set of
simulations is created by sampling the model parameters according to their
\emph{a priori} distributions. If there are stochastic processes, these
simulations are repeated for each parameter value to sample the variation in
QOI created by the intrinsic stochasticity. The distribution of the QOI can
then be reconstructed using standard kernel density methods \cite{izenman1991review}.
However, in large-scale simulations, this type of Monte Carlo approach is
prohibitively expensive. In this case, an \emph{emulator} or \emph{statistical
  surrogate model} can be used to characterize the QOIs over a range of
possible model parameters \cite{higdon2008computer,oakley2002bayesian,o1998uncertainty,sargsyan2010spectral}.
Sufficient samples are generated until this statistical process faithfully
reproduces the observed correlations and can account for uncertainty due to
finite sampling of the simulation. These processes are then used as surrogate
models to quantify the model uncertainty and the correlations of the QOIs to
the input parameter values.

We refer to uncertainty (variation) in QOI due to imprecisely known input
parameters as \emph{parametric} or \emph{epistemic} uncertainty
\cite{helton2010representation}. With parametric uncertainty, we do not know
the specific model parametrization needed in our simulation to make accurate
predictions. It is assumed that a believable range of values for a parameter
and the probability associated with a value in that range is known, i.e., we
know the \emph{probability density function} (pdf) of the input
parameter(s). Examples of parametric uncertainty abound; in epidemiology, the
mean and variance of recovery rates for diseases are typically determined
experimentally, whereas transmission rates are typically obtained from
observing epidemic progression in a population. Parametric uncertainty
represents uncertainty in QOI due to imprecisely known input parameters to the
simulation.

In addition to parametric uncertainty, some models' predictions rely on the
outcome of random events, which create uncertainty in the model predictions,
even if input parameters are fixed. We refer to these stochastic variations in
model QOI as \emph{intrinsic} or \emph{aleatory} uncertainty
\cite{helton2010representation}.  This type of uncertainty is observed in
epidemic models when the number of individuals becoming infected on a
particular day is a random event and a function of the stochastic nature of a
communities' contact network structure
\cite{allen2008introduction,newman2002spread}. Intrinsic uncertainty
represents variation in QOI that is present even when input parameters for the
simulation are fixed.

The two types of uncertainty can be closely connected. For a specific example,
the mean and variance of the distribution for the time it takes a person to
recover from a disease may be specified as inputs to an epidemic simulation
but may only be known imprecisely. If the imprecise knowledge is specified by
a known probability density function we would label this mean and variance as
a source of parametric uncertainty. Once the mean and variance are fixed,
however, each simulation of a stochastic epidemic model will result in a
distinct epidemic realization. The variation in these realizations, with the
input parameters fixed, is labeled as a source of intrinsic uncertainty
\cite{allen2008introduction,allen2000comparison}.

Our UQ approach combines multiple methods of statistical surrogate modeling to
allow separation of parametric and intrinsic uncertainty. We show how to
construct a statistical emulator that distinguishes between the two types of
uncertainty and can account for situations where the intrinsic uncertainty is
non-Gaussian. In the presence of intrinsic uncertainty, the simulations are
sampled randomly for each fixed input parameter. A kernel density approach
\cite{izenman1991review} is used to estimate the distribution of QOI for each fixed
parameterization, and the contribution to the variation from intrinsic sources
is separated using a non-intrusive polynomial chaos method
\cite{sargsyan2010spectral}. The inclusion of the polynomial chaos
decomposition allows our method to account for non-Gaussianity in the
intrinsic variation of the QOI. Once the polynomial chaos decomposition has
been performed, the contribution to variation from parametric uncertainty can
be analyzed separately using a Gaussian process emulator
\cite{higdon2008computer,oakley2002bayesian,o1998uncertainty}.

Since the emulator adds additional uncertainty when it interpolates QOI in
regions where there are few samples, the surrogate model constructed here has
a variance that increases at parameter values far from samples of the
simulation.  We will describe the approach for a situation where the QOIs can
be approximated by a unimodal distribution. If this condition is not
satisfied, then clustering methods can be used to reduce the QOI distribution
to several unimodal random variables and apply the emulator to each one
separately \cite{sargsyan2010spectral}. In this work, we eliminate the
multi-modality of a simulations' output in a pre-processing step by
considering as output, simulation results that fall close to one of the modes
in the distribution of simulation predictions. This step can be thought of as
studying the random variable that is the QOI from the simulation, conditioned
on the event that the prediction is near a particular mode. Each mode can then
be studied as a separate conditional response of the simulation.

Although our approach uses fewer samples than a standard Monte Carlo method,
it can still require significant computational resources to construct the
surrogate model, especially if there are many parameters with intrinsic or
parametric uncertainties. The emulator itself is an approximation and samples
from the emulation will not exactly reproduce the distributions of the models
of QOI. Although the mean and variance of the emulated probability
distributions may converge rather quickly, the higher order moments can
require a large sample size. As more samples of the simulation are included in
the emulator construction, the emulation will behave more like the actual
simulated model, though the exact rate of convergence remains an open
question.

In the next section, we introduce the stochastic epidemic simulation that we
use as an example throughout the paper. We describe how to account for
intrinsic uncertainty using the Karhunen-Loeve decomposition and a
non-intrusive polynomial chaos decomposition. This is followed by our
implementation of Gaussian process regression to model the effect of
parametric uncertainty and finite sample size.

\section{The stochastic epidemic model}
 
Throughout the description of our surrogate modeling methodology, we will keep
in mind a particular example coming from epidemic modeling
\cite{allen2008introduction,allen2000comparison}. This work was motivated by
the lack of approaches to quantify uncertainty for large scale agent-based
epidemic models \cite{grefenstette2013fred,mniszewski2008episims}. Our
emulation method is general enough to be applied to any mathematical model
simulation but due to our original motivation, it will be demonstrated using a
stochastic epidemic model. The model is a system of stochastic differential
equations (SDE). Each term in the SDE, represents the number of individuals in
a particular disease state, i.e., an individual is \emph{susceptible} to the
disease, \emph{infected} with the disease and can infect others, or
\emph{recovered} from the disease and has immunity. In the epidemic modeling
literature, this is known as the classical
\emph{Susceptible-Infected-Recovered} (SIR) disease model
\cite{anderson1991infectious}.

The differential equations for our SIR model are briefly outlined here, for
further reference and a complete derivation see
\cite{allen2008introduction,allen2000comparison}. The constant size of the
population is denoted $N$. The number of individuals in each category at time
$t > 0$ is denoted as $\mathcal{S}(t)$, $\mathcal{I}(t)$, $\mathcal{R}(t)$ for
susceptible, infected, and recovered, respectively. Since the total population
is constant, the number of recovered individuals satisfies $\mathcal{R}(t) = N
- \mathcal{S}(t) - \mathcal{I}(t)$. Letting $Z(t) = (\mathcal{S}(t),
\mathcal{I}(t))^T$ and defining the time dependent mean and covariance
\begin{equation}
A(Z) = \left( \begin{array}{c}
-\frac{\beta}{N} \mathcal{S} \mathcal{I} \\
\\
\frac{\beta}{N} \mathcal{S} \mathcal{I} - \gamma \mathcal{I}
\end{array} \right),\,
V(Z) = \left( \begin{array}{cc}
\frac{\beta}{N} \mathcal{S} \mathcal{I} & -\frac{\beta}{N} \mathcal{S} \mathcal{I} \\
& \\
-\frac{\beta}{N} \mathcal{S} \mathcal{I} & \frac{\beta}{N} \mathcal{S} \mathcal{I} + \gamma \mathcal{I}
\end{array} \right)
\end{equation} 
we can express the time evolution of $Z(t)$ by the It$\hat{\textrm{o}}$ SDE
\cite{karatzas1991brownian}
\begin{equation}\label{ito_SIR}
dZ(t) = A(Z) dt + B(Z) d\mathcal{W},\,\, Z(0) = (\mathcal{S}_0, \mathcal{I}_0)^T.
\end{equation}
Here $B(Z) = \sqrt{V(Z)}$, $V(Z)$ is symmetric and positive definite, and
$\mathcal{W} = (\mathcal{W}_1, \mathcal{W}_2)^T$ is a vector of independent
Wiener processes \cite{karatzas1991brownian}. The parameter $\beta > 0$ is the
infection rate, representing the rate at which individuals in the population
become infected. The parameter $\gamma > 0$ is the recovery rate, representing
the rate at which an infected individual is removed from the population. In
this model, once an individual recovers, they are considered either dead or
immune, and are removed from the model completely.

\begin{figure}[ht]
  $\begin{array}{cc}
      \includegraphics[scale=0.3]{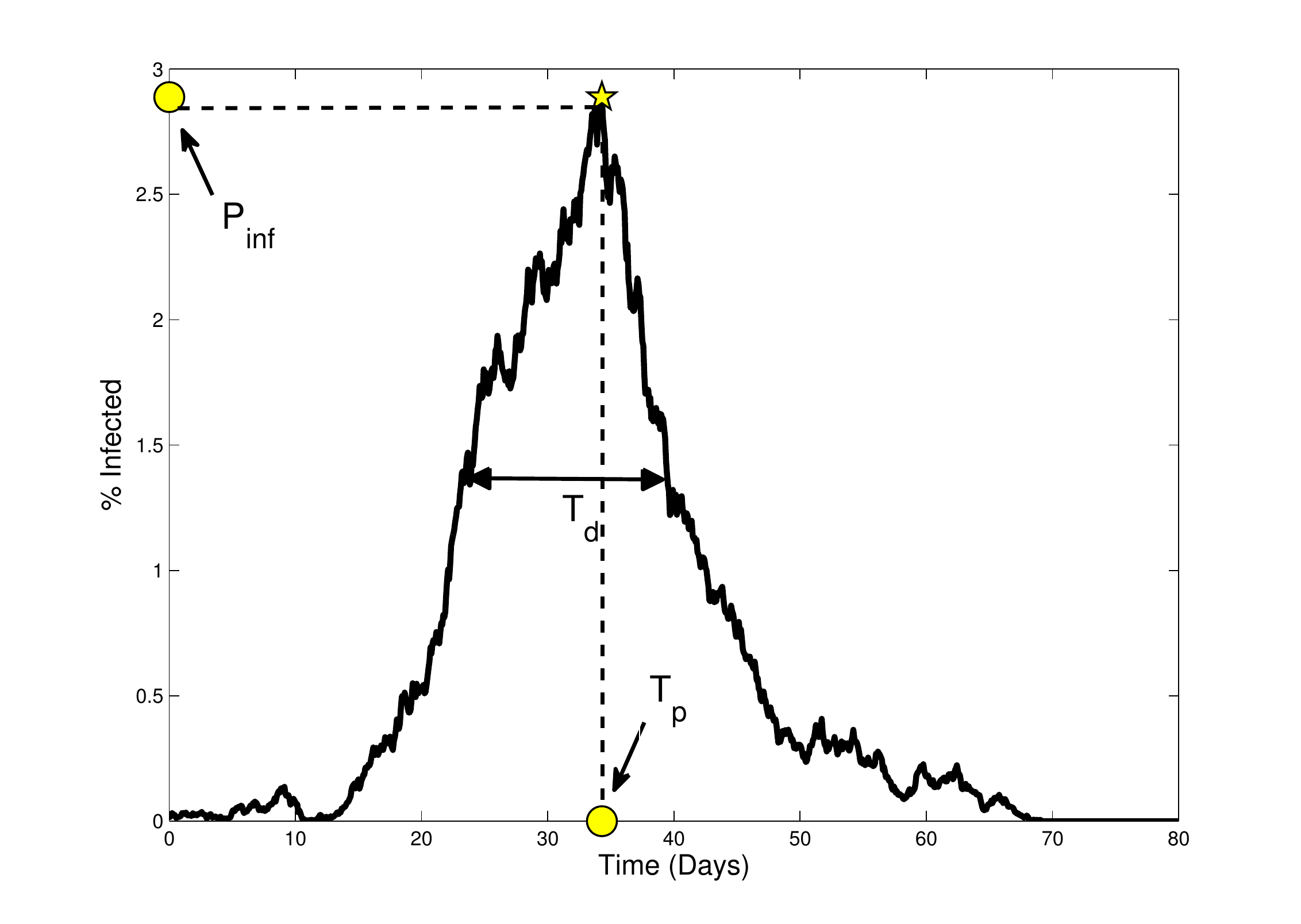} &
      \includegraphics[scale=0.3]{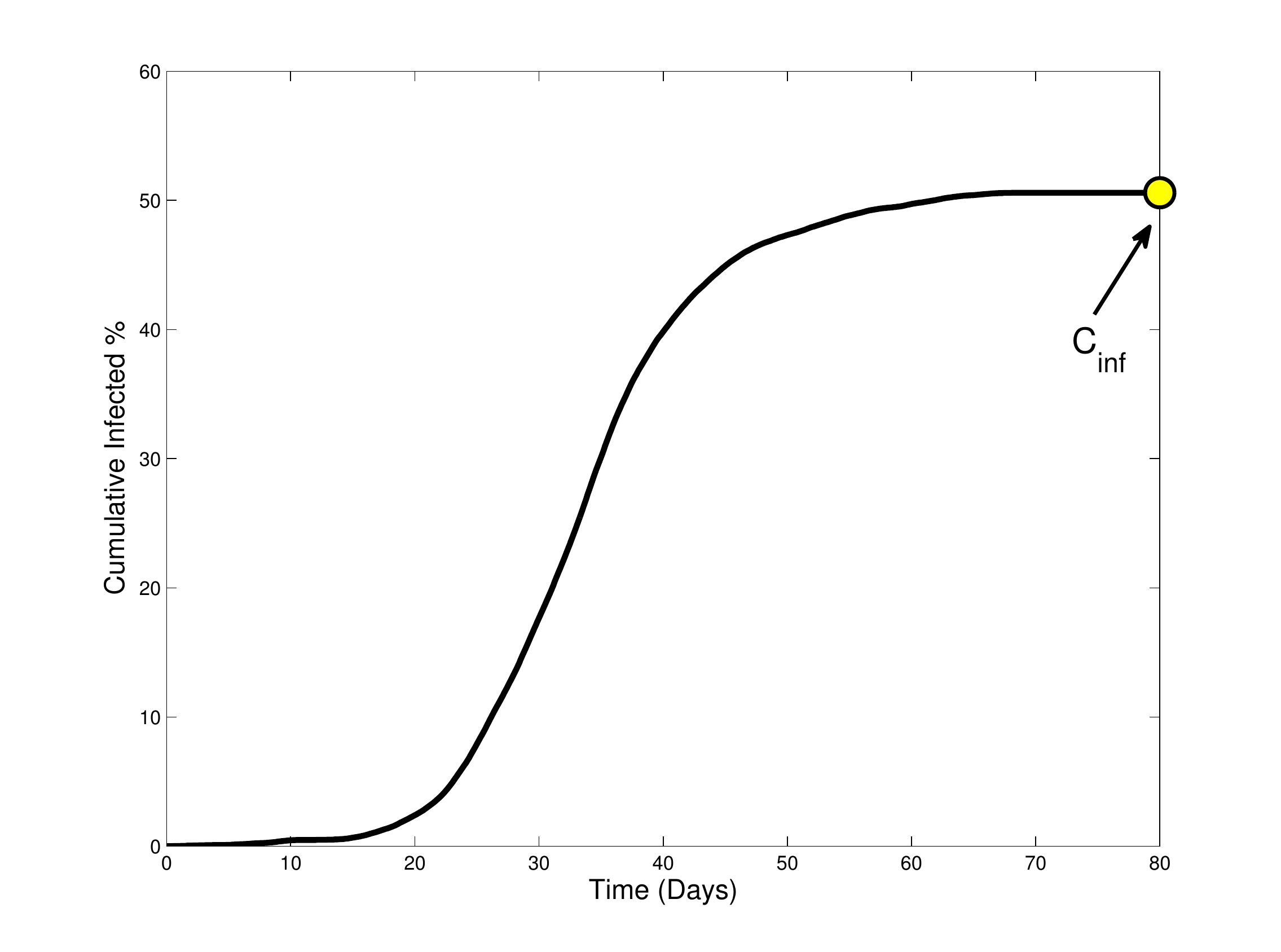} \\
    \end{array}$
    \caption{A sample of $\vec{Q}(\beta, \gamma; \omega)$ from one SIR
      realization. (LEFT) \% infected time series for a fixed $(\beta,
      \gamma)$. $Q_1$, $Q_2$, and $Q_3$ are labeled. (RIGHT) \% cumulatively
      infected time series for a fixed $(\beta, \gamma)$, $Q_4$ is marked.}
\label{fig:SIR_QOI}
\end{figure}

In our example, for a given simulation, we analyze a population of $N =
10,000$ individuals with initial conditions $Z(0) = (9998, 2)^T$. When a
recovery rate and infection rate $(\beta, \gamma)$ are fixed and a simulation
is run, we record the following four QOI:
\begin{equation}
\begin{aligned}
Q_1(\beta, \gamma; \omega) &:= P_{\textrm{inf}} = \textrm{Maximum \% population simultaneously infected} \\
Q_2(\beta, \gamma; \omega) &:= T_p = \textrm{Time to the peak}, P_{\textrm{inf}} \textrm{, in days} \\
Q_3(\beta, \gamma; \omega) &:= T_d = \textrm{Duration, number of days \% infected is within 50\% of peak} \\
Q_4(\beta, \gamma; \omega) &:= C_{\textrm{inf}} = \textrm{Cumulative \% ever infected}
\end{aligned}
\end{equation}
produced by the simulation. These four QOI are depicted for one sample of the
stochastic SIR simulation in Figure \ref{fig:SIR_QOI}. The state variable
$\omega$ indicates random variation due to the stochastic effects in the
SDE. After the simulated data is generated, we restrict ourselves to studying
only the simulations where at least $10\%$ of the population becomes
infected. This will have the effect of making the joint distribution of the
QOI, $(Q_1, Q_2, Q_3, Q_4)$, approximately unimodal, which is a necessary
assumption for our emulation method. The vector output of the simulation will
be denoted by
\begin{equation}
\vec{Q}(\beta, \gamma; \omega) = (Q_1, Q_2, Q_3, Q_4)^T.
\end{equation}

\begin{figure}[h]
  \begin{center}
      \includegraphics[scale=.6]{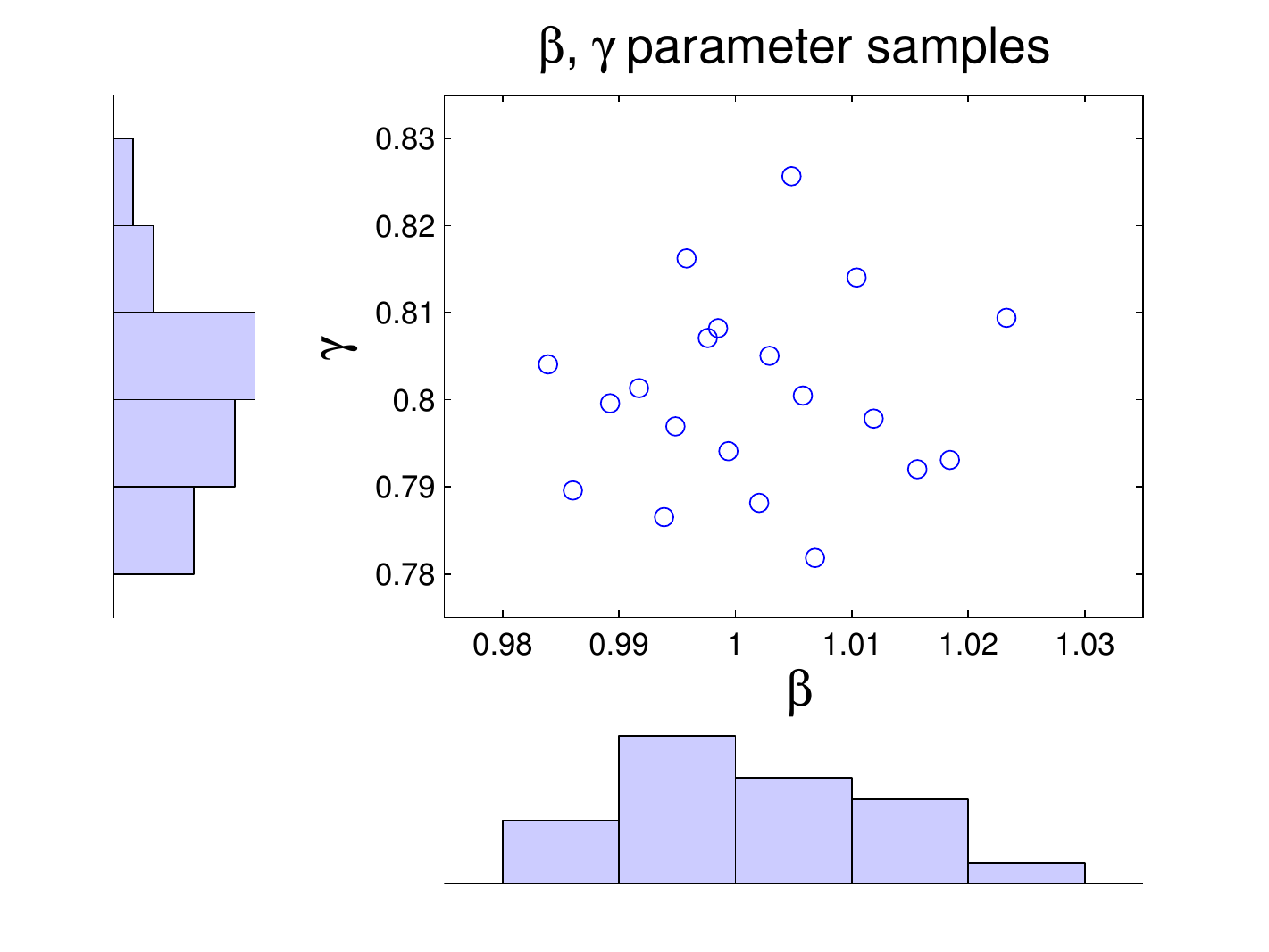}
  \end{center}
  \caption{Samples of $(\beta, \gamma)$ parameters from their log-normal
    distributions and corresponding univariate histograms of the
    samples. These samples are then used as input parameters to
    (\ref{ito_SIR}), effectively exploring the contribution of parametric
    uncertainty to the distribution of $\vec{Q}(\beta, \gamma; \omega)$.}
  \label{fig:param_samples}
\end{figure}

\begin{figure}[h]
  \begin{center}
      \includegraphics[scale=.4]{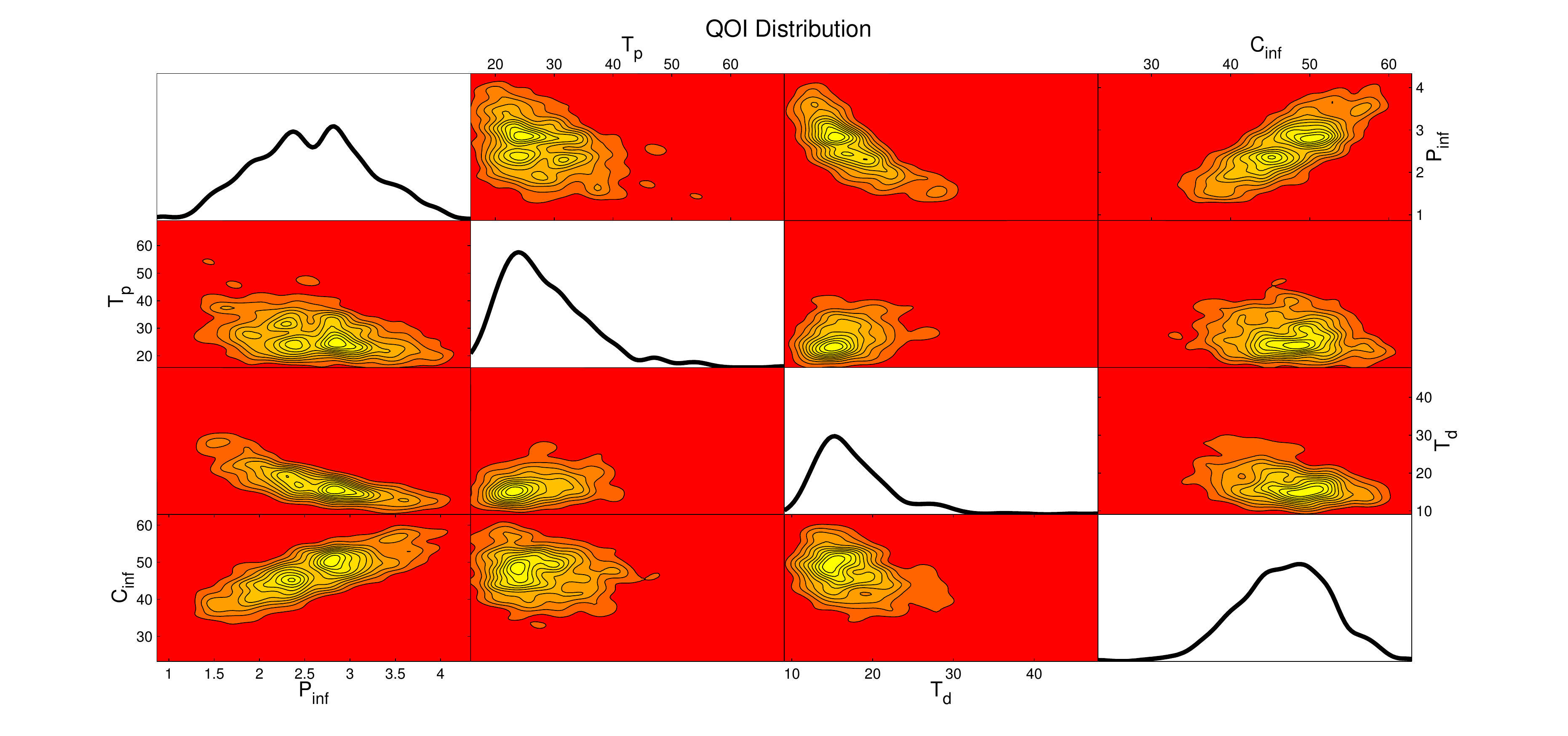}
  \end{center}
  \caption{One dimensional and two dimensional marginal distributions from the
    four dimension distribution corresponding to $\vec{Q}(\beta, \gamma;
    \omega)$. Kernel density estimation was used to generate each of the
    marginal distributions from samples of the SIR model. The goal of the
    methods detailed in this paper is to be able to reconstruct an
    approximation of this distribution from very few samples of the actual SIR
    simulation and to be able to directly generate approximate samples of
    $\vec{Q}(\beta, \gamma; \omega)$ quickly.}
  \label{fig:QOI_trueKDE}
\end{figure}

The transmission and recovery rate parameters, $(\beta, \gamma)$, are our
source of parametric uncertainty. In practice, these are determined from
experimental data or observations of disease spread in similar
populations. They are only approximately determined and we model their
uncertainty by treating them as random variables. We assume that $(\beta,
\gamma)$ have lognormal distributions,
\begin{equation}\label{param_uncertainty}
\beta \sim \ln \Norm(\mu_{\beta}, \sigma^2_{\beta}),\,\, \gamma \sim \ln \Norm(\mu_{\gamma}, \sigma^2_{\gamma}), 
\end{equation}
with $\mu_{\beta} = 1$, $\sigma^2_{\beta} = 0.000125$, $\mu_{\gamma} = 0.8$,
$\sigma^2_{\gamma} = 0.000125$. These distributions and their corresponding
univariate histograms of the samples are plotted in Figure
\ref{fig:param_samples}. Besides uncertainty in the QOI caused by imprecisely
known transmission and recovery rates, we must also account for the variation
introduced by the stochasticity in the SIR model, an intrinsic
uncertainty. When sampling $\vec{Q}(\beta, \gamma; \omega)$ to quantify the
model's uncertainty, we first draw $(\beta, \gamma)$ samples from the
distributions in Equation (\ref{param_uncertainty}). For each of these
parameter sets, we then simulate multiple solutions to Equation
(\ref{ito_SIR}) and record the samples of $\vec{Q}(\beta, \gamma;
\omega)$. Repeating the simulation multiple times for each parameter set is
essential to sample the contribution of intrinsic uncertainty to the
distribution of the QOI. Once this is done, we use \emph{kernel density
  estimation} \cite{izenman1991review} to form an approximate distribution of
$\vec{Q}(\beta, \gamma; \omega)$. The result of this process is shown in
Figure \ref{fig:QOI_trueKDE}.

\paragraph{Notation:}To ensure generality in our presentation and to avoid
overly cumbersome notation, we denote $X(\theta; \omega) = (X_{\tau}(\theta;
\omega))_{\tau=1}^d \in \R^d$ to represent our vector quantity of
interest. Here, $\tau$ is a discrete index parameter indicating the specific
QOI. In practice, $\tau$ could also represent discrete samples of a continuous
parameter, e.g., time. The vector $\theta \in \R^p$ will denote input
parameters to the simulation and $\omega$ is the state variable controlling
stochastic variation.

For the simulation, output $X_{\tau}(\theta; \omega)$ uncertainty due to
variation in $\theta$ will be \emph{parametric uncertainty}. We will assume
that we know the probability density for $\theta$ and can choose how we will
sample from that density when the simulation is run. The uncertainty in our
output due to $\omega$ will be referred to as \emph{intrinsic uncertainty},
which is characterized by the absence of a parameterized probability space and
by our inability to choose how to sample its effects.

For the SIR model, the notational translation will be $X_{\tau}(\theta;
\omega) = Q_{\tau}(\beta, \gamma; \omega)$ with $\tau = 1,2,3,4$ and $\theta =
(\beta, \gamma)$.

\section{Karhunen-Loeve decomposition}

In this section, we use the Karhunen-Loeve (KL) decomposition to transform the
QOI to a set of uncorrelated random variables. The lack of correlation between
the transformed QOI will aid in our construction of a polynomial chaos
representation of our model. Using the KL decomposition has the added benefit
of possibly reducing the dimension of the QOI space being emulated.

We first decompose $X_{\tau}(\theta; \omega)$ so that the contributions from
$\tau$ and $\theta$ are separated from those of $\omega$. Define
\begin{equation*}
  \overline{X} = \E[X(\theta; \omega)]
\end{equation*}
so the process may be represented as 
\begin{equation}
  X(\theta; \omega) = \overline{X} + X^0(\theta; \omega).
\end{equation}
The zero mean random vector $X^0(\theta; \omega)$ is now emulated.

In order to remove the correlations between the QOI and reduce dimensionality,
we use a Karhunen-Loeve (KL) \cite{karhunen1947lineare,loeve1978probability}
decomposition for $X^0$. The covariance function is given by the $d \times d$
square matrix
\begin{equation}
  K(s,r) = \E[X^0_s(\theta; \omega) X^0_r(\theta; \omega)], \,\, s, r = 1,2, \dots, d.
\end{equation}
In the KL decomposition, one then finds the eigenfunctions of the covariance
by solving the eigenvalue problem
\begin{equation}\label{covariance_eigs}
  \sum_{r=1}^d K(\tau,r) f^n_r = \lambda_n f^n_{\tau}, \,\, \tau, n = 1,2, \dots, d.
\end{equation}
The set of Euclidean unit length eigenvectors $\{ f^n \}_{n=1}^{d} \subset
\mathbb{R}^d$ forms a basis for the random vector $X^0(\theta; \omega)$ so we
may project onto this basis. Coefficients of the projection are
\begin{equation}
  \xi_n(\theta; \omega) = \langle X^0(\theta; \omega), f^n \rangle = \sum_{\tau = 1}^d X^0_{\tau}(\theta; \omega) f^n_{\tau}. 
\end{equation}
The KL decomposition of the zero mean random vector is
\begin{equation}
  X^0(\theta; \omega) = \sum_{n=1}^{d} \xi_n(\theta; \omega) f^n
\end{equation}
and the \emph{truncated} KL decomposition is obtained by taking the first $N <
d$ terms in the series
\begin{equation}\label{KLdecomposition}
  X^0(\theta; \omega) \approx \sum_{n=1}^N \xi_n(\theta; \omega) f^n.
\end{equation}

With Equation (\ref{KLdecomposition}), we have reduced the emulation problem
to the problem of emulating the random vector of \emph{uncorrelated}
coefficients $\xi(\theta; \omega) =
(\xi_1(\theta;\omega),\dots,\xi_N(\theta;\omega))$. Since we are not assuming
that $X(\theta; \omega)$ is Gaussian the coefficients, $\xi_i(\theta;
\omega)$, are not necessarily Gaussian and therefore not necessarily
independent.

Correlations between entries of $X(\theta; \omega)$, corresponding to
different QOI, is now controlled by the entries of the eigenvectors $f^n$. If
the eigenvalues in Equation (\ref{covariance_eigs}) decay rapidly, the number
of terms, $N$, can be taken to be much smaller than the original number of
QOI, effectively reducing the output dimension.

\begin{figure}[h]
  \begin{center}
      \includegraphics[scale=.4]{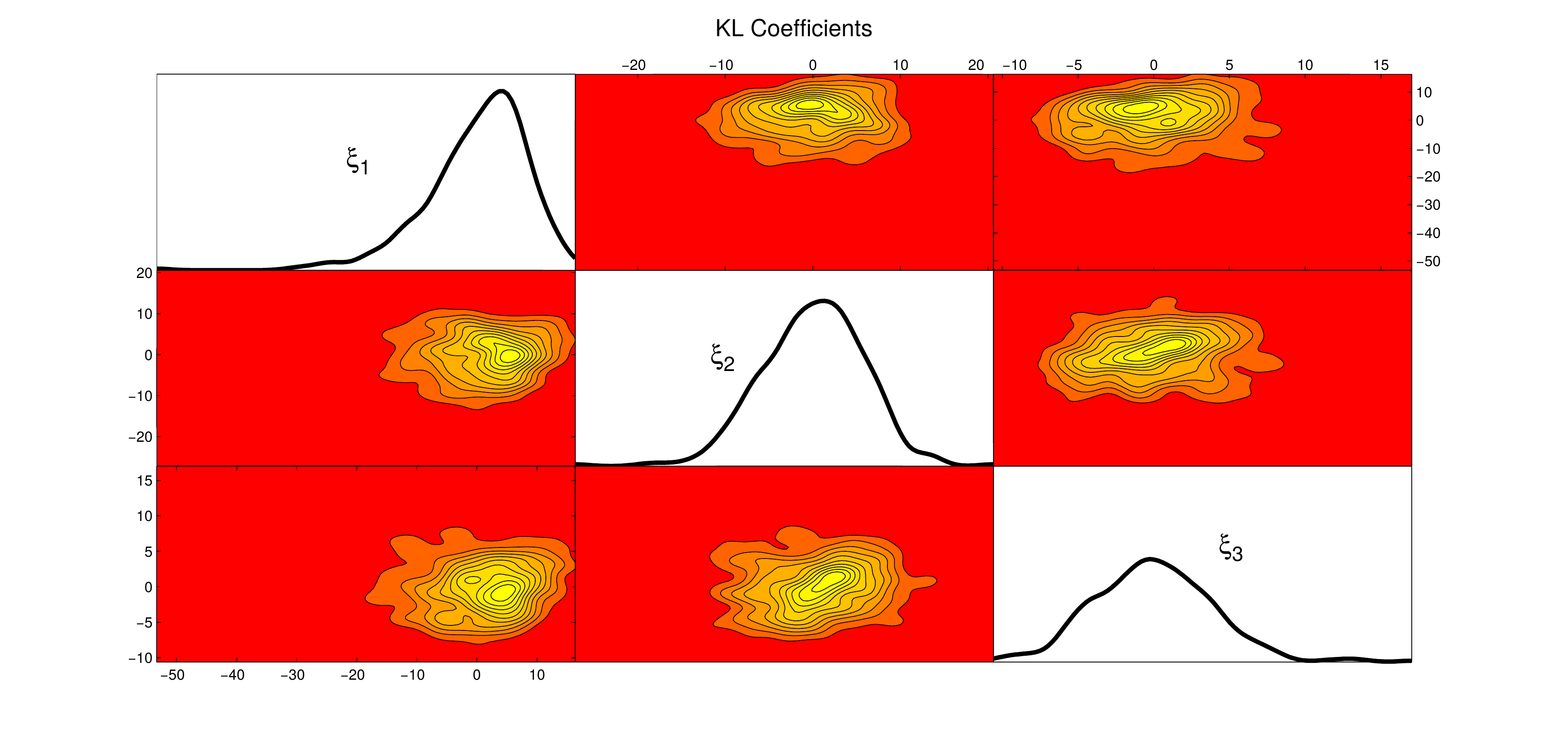}
  \end{center}
  \caption{Starting with the four dimensional distribution of $\vec{Q}(\beta,
    \gamma; \omega)$, shown in Figure \ref{fig:QOI_trueKDE}, the KL
    decomposition is computed. In this figure, we show the one and two
    dimensional marginal distributions for the first three KL coefficients
    derived from the distribution of $\vec{Q}(\beta, \gamma; \omega)$. When
    compared to the marginal distributions in Figure \ref{fig:QOI_trueKDE},
    one can see that the correlations between the KL coefficients is less than
    those of the original QOI. This is important for implementation of the
    polynomial chaos decomposition.}
  \label{fig:KLcoeff}
\end{figure} 

\begin{figure}[h]
  \begin{center}
      \includegraphics[scale=.4]{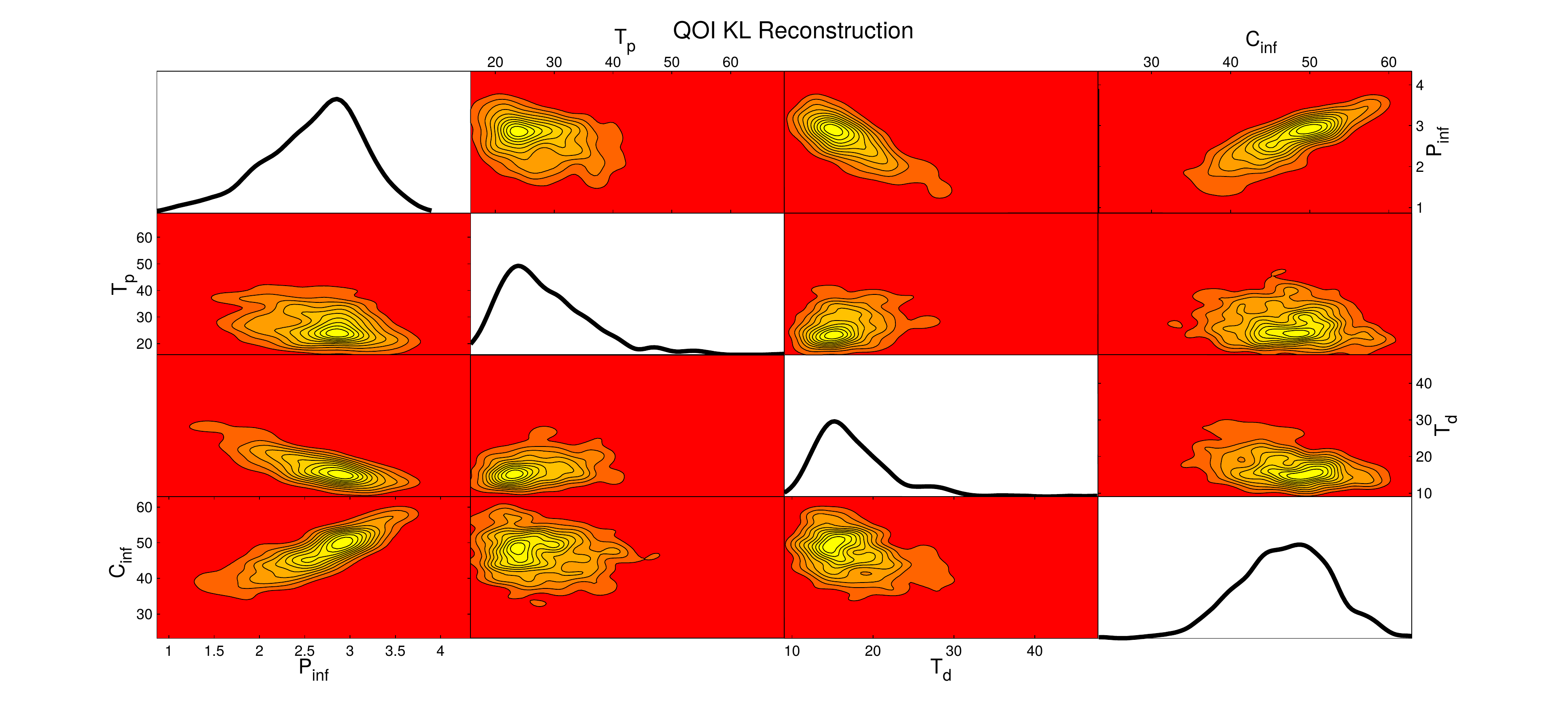}
  \end{center}
  \caption{The first three KL coefficients depicted in Figure
    \ref{fig:KLcoeff} are used to reconstruct an approximation to the
    distribution of $\vec{Q}(\beta, \gamma; \omega)$. Using this approximation
    allows us to reduce the dimension of the QOI.}
  \label{fig:KLrecon_QOI}
\end{figure} 

The effectiveness of the KL decomposition on our example problem is shown in
Figures \ref{fig:KLcoeff} and \ref{fig:KLrecon_QOI}. Since our SIR model
output, $\vec{Q}(\beta, \gamma; \omega)$, is only four dimensional, a complete
KL decomposition has four coefficients. In Figure \ref{fig:KLcoeff}, the
distribution of the first three KL coefficients is visualized. These
coefficients are used to reconstruct an approximation to the distribution of
$\vec{Q}(\beta, \gamma; \omega)$ shown in Figure \ref{fig:QOI_trueKDE}. The
approximate reconstruction has a distribution depicted in Figure
\ref{fig:KLrecon_QOI}.

\section{Polynomial chaos expansion of coefficients}

A reduced representation for the random vector of uncorrelated, but not
independent, random variables $\xi(\theta; \omega) =
(\xi_1(\theta;\omega),\dots,\xi_N(\theta;\omega))$ is constructed using a
\emph{polynomial chaos} (PC) decomposition. This accomplishes two
goals. First, at a fixed $\theta$, the PC decomposition computes a low
dimensional approximation to the distribution of $\xi(\theta; \omega)$ while
still allowing for non-Gaussianity. Second, the PC approximation gives us a
way to generate approximate samples from the distribution of $\xi(\theta;
\omega)$ at a fixed parameterization $\theta$. Although KL gives a low
dimensional approximation to the orginal distribution of $X(\theta; \omega)$
it does not provide a way to generate samples from that distribution.

In polynomial chaos, a random variable is approximated by a series of
polynomial basis functions that form a basis for the underlying probability
space of the random variable. It has been shown \cite{xiu2003wiener} that the
particular choice of basis makes a significant difference in the rate of
convergence to the random variable. We will present our methods using the
Hermite polynomials \cite{xiu2003wiener}, which work well for our SIR example,
though the techniques used are independent of the basis. Different basis of
polynomials, in the generalized polynomial chaos scheme, are orthogonal with
respect to different measures. The Hermite polynomials are orthogonal with
respect to the standard Gaussian density. This makes them ideal for
application when the underlying distribution of the random variable being
emulated is approximately normal. If there is reason to suspect that a
distribution is far from normal then another basis should be considered
\cite{xiu2003wiener}.

The Hermite polynomial of degree $k$ in a single dimension is denoted by
$\psi_k(x)$. This set of polynomials can be defined recursively by
\begin{align}
\psi_0(x) &= 1 \\ \nonumber
\psi_1(x) &= x \\ \nonumber
\psi_{m+1}(x) &= x \psi_m(x) - m \psi_{m-1}(x) \textrm{ for } m = 1, 2, \dots.
\end{align}
Since the Hermite polynomials are orthogonal with respect to the standard
normal density
\begin{equation*}
  w(x)\,dx = \frac{1}{\sqrt{2 \pi}} e^{-x^2/2}\, dx,
\end{equation*} 
the relation 
\begin{equation*}
\frac{1}{\sqrt{2 \pi}} \int_{-\infty}^{\infty} \psi_i(x) \psi_j(x) e^{-x^2/2}\, dx = j! \delta_{ij}
\end{equation*}
holds for all $i,j = 0,1,2,\dots$. The Hermite polynomials in higher
dimensions are formed by tensor product. Let $\alpha = (\alpha_1, \alpha_2,
\dots , \alpha_d) \in \Z^d$ be a multi-index and define the Hermite polynomial
on $\R^d$ by
\begin{equation}
\Psi_{\alpha}({\bf x}) = \psi_{\alpha_1}(x_1) \psi_{\alpha_2}(x_2) \dots \psi_{\alpha_d}(x_d).
\end{equation}
To index the Hermite polynomials in higher dimensions, we use a \emph{graded
  lexicographic ordering} \cite{cox2007ideals} on the multi-indices in
dimension $d$. That is, we will use $\Psi_k({\bf x})$ to refer to the
multidimensional Hermite polynomial with the $k^{th}$ multi-index in the
graded lexicographic ordering.

\begin{figure}[h]
  \begin{center}
      \includegraphics[scale=.4]{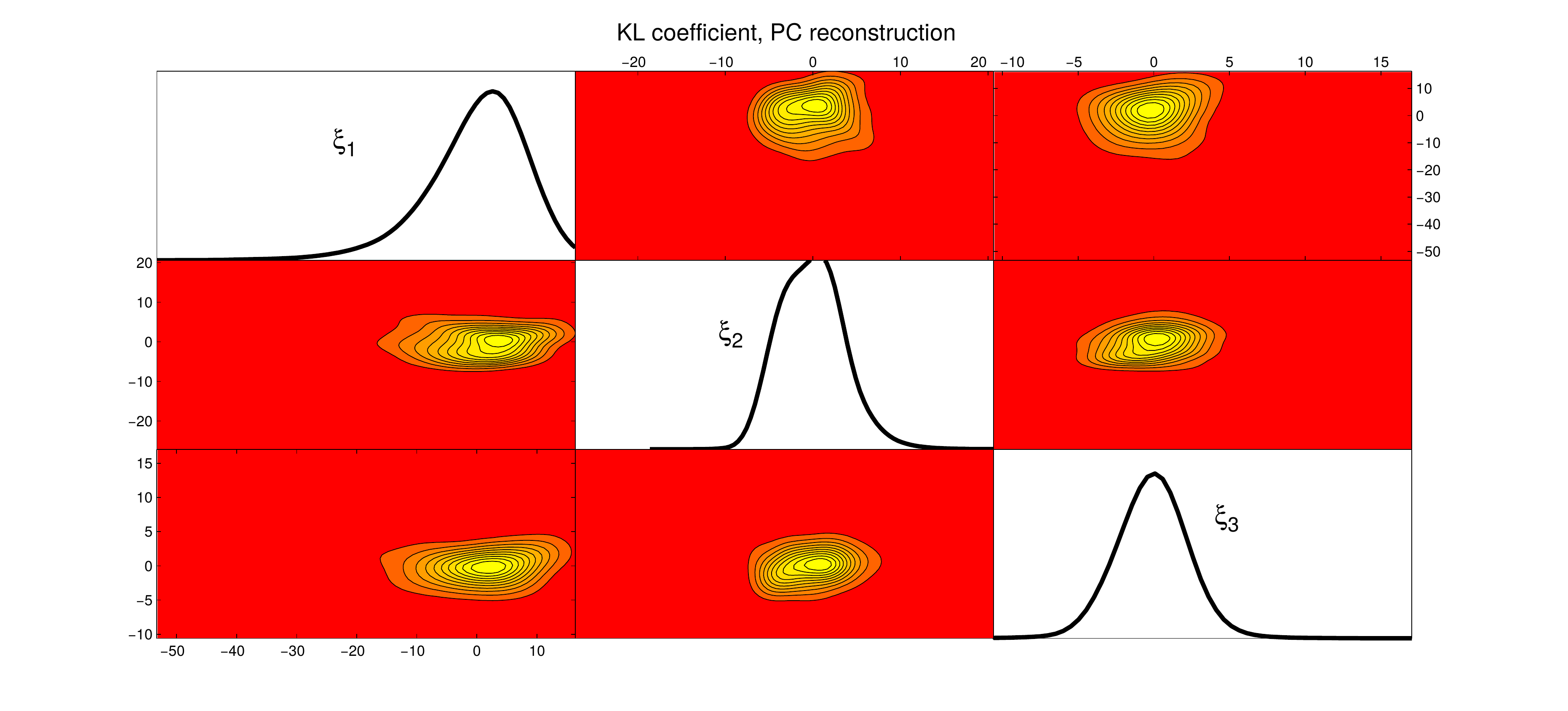}
  \end{center}
  \caption{Reconstruction of the one and two dimensional marginal
    distributions for KL coefficients using the first eight terms in the
    polynomial chaos reconstruction given by Equations
    (\ref{PCexpansion})--(\ref{PCexpansion_indexed}). Since the Hermite
    polynomial chaos approximation is formed from smooth functions sampling
    from the truncated decomposition has the effect of smoothing the original
    distribution and clipping the low probability regions. This can be
    observed by comparing the above Figure with the original KL distribution
    in Figure \ref{fig:KLcoeff}.}
  \label{fig:KLcoefPCrecon}
\end{figure}

With the above indexing on higher dimensional Hermite polynomials, the
polynomial chaos decomposition of the random vector $\xi(\theta; \omega)$, up
to order $K$, is given by
\begin{equation}\label{PCexpansion}
\xi(\theta)  = \xi(\theta; \omega(\zeta)) \approx \sum_{k=0}^K {\bf c}_k(\theta) \Psi_k(\zeta).
\end{equation}
Here ${\bf c}_k(\theta)$ is a length $N$ vector of coefficients corresponding
to each entry in $\xi$ so, for $n = 1,2,\dots,N$,
\begin{equation}\label{PCexpansion_indexed}
\xi_n(\theta) = \xi_n(\theta; \omega(\zeta)) \approx \sum_{k=0}^K c_{nk}(\theta) \Psi_k(\zeta).
\end{equation}
The new variable $\zeta = (\zeta_1, \zeta_2, \dots, \zeta_N)$ is a random
vector of independent standard normal random variables. This vector serves the
purpose of parameterizing the probability space corresponding to
$\omega$. That is, sampling from $\zeta$ is equivalent to sampling from
$\omega$. In Figure \ref{fig:KLcoefPCrecon}, we show the reconstruction of the
first three KL coefficients for the QOI in the stochastic SIR model using
Equation (\ref{PCexpansion_indexed}).

The difficulty with the expansion given by Equations
(\ref{PCexpansion})--(\ref{PCexpansion_indexed}) is the computation of the
coefficients ${\bf c}_k(\theta)$. These are formally defined through Galerkin
projection onto the Hermite polynomials \cite{xiu2003wiener} by the formula
\begin{equation}\label{PCcoefproj}
c_{nk}(\theta) = \frac{\E[\xi_n(\theta;\omega) \Psi_k(\zeta)]}{\E[\Psi_k^2(\zeta)]}.
\end{equation} 
This formula is only formal since $\xi_n(\theta;\omega)$ and $\Psi_k(\zeta)$
live over different probability spaces. Computing the expectation
$\E[\xi_n(\theta;\omega) \Psi_k(\zeta)]$ in Equation (\ref{PCcoefproj}) is
performed by Monte Carlo approximation, so sampling of $\xi_n(\theta;\omega)$
and $\Psi_k(\zeta)$ must take place over the same probability space. To
compute this expectation, we transform the two probability spaces to a common
domain. There is a standard method to transform two different, finite
dimensional, probability spaces into a common space, which usually relies on
having an explicit representation of the underlying distributions. Lacking an
explicit representation, we will form an approximation using a kernel density
estimate.

The Rosenblatt transformation \cite{rosenblatt1952remarks} uses the
conditional cumulative distribution functions to map a set of jointly
distributed random variables onto a set of independent uniform random
variables on $[0,1]$. In terms of the conditional cumulative distribution, the
Rosenblatt transformation is given by
\begin{align}\label{rosenblatt}
u_1 &= F_1(\xi_1) \\ \nonumber
u_2 &= F_{2|1}(\xi_2 | \xi_1) \\ \nonumber
u_3 &= F_{3|1,2}(\xi_3 | \xi_1, \xi_2) \\ \nonumber
&\vdots \\ \nonumber
u_N &= F_{N|1,2,\dots,N-1}(\xi_N | \xi_1, \xi_2, \dots, \xi_N). \nonumber
\end{align}
Once this map is computed, the cumulative distributions are used to generate
samples from the joint distribution of the random vector $\xi$ from $N$
independent samples of uniform random variables on $[0,1]$. This process is
the inverse Rosenblatt transformation, which maps a set of $N$ independent
uniformly distributed random variables to the random vector $\xi$ of length
$N$. This process uses the inverses of each of the marginal cumulative
distributions in Equation (\ref{rosenblatt}). We will denote this inverse
Rosenblatt transform by
\begin{equation}
g({\bf u}) = \xi, {\bf u} \sim U[0,1]^N.
\end{equation}
Likewise it is standard to map ${\bf u} \sim U[0,1]^N$ to independent normally
distributed random variables $\zeta \sim N(0,1)^N$. We denote this map by
\begin{equation}
l({\bf u}) = \zeta, {\bf u} \sim U[0,1]^N.
\end{equation}

With the above maps one can then compute the expectation in the numerator of
(\ref{PCcoefproj}) as follows,
\begin{align}\label{PCcoef_transformed}
\E[\xi_n(\theta;\omega) \Psi_k(\zeta)] &= \frac{1}{(2 \pi)^{N/2}}\int \xi_n(\theta;\omega) \Psi_k(\zeta)
e^{-\frac{1}{2}\|\zeta\|^2}\, d\zeta \\ \nonumber
&= \int_{[0,1]^N} g_{\theta}({\bf u}) \Psi_k(l({\bf u})) \,d{\bf u}. \nonumber
\end{align}
We note that the integral is computed using a Monte Carlo approach and we have
used the notation $g_{\theta}({\bf u})$ to indicate that the particular
inverse Rosenblatt transformation depends on the parameter $\theta$. As stated
above, to calculate the expectation in Equation (\ref{PCcoef_transformed})
using a Monte Carlo scheme one must be sure to use the same sample of the
uniform random variable ${\bf u}$ when calculating values for $g_{\theta}({\bf
  u})$ and $l({\bf u})$. It is also important to use a Monte Carlo scheme of
sufficient order relative to the size of the coefficients involved in Equation
(\ref{PCexpansion}).

When using Equation (\ref{PCexpansion}) to characterize intrinsic uncertainty
in a simulation, one does not have an explicit form for the conditional
cumulative functions in (\ref{rosenblatt}). In practice, for a fixed $\theta$,
one only has a finite number of samples of the random vector
$\xi(\theta;\omega)$, which we will denote by $\{\xi^{(m)}\}_{m=1}^M$. To
estimate the conditional cumulative functions, we first use the samples to
estimate the joint pdf of the random vector $\xi$. This can be done using a
\emph{kernel density estimation} (KDE) method \cite{izenman1991review}. We use
univariate tensor product kernels where our univariate kernel for the $i^{th}$
variable is denoted by $K_i(\xi_i)$ and our bandwidth is denoted by $h >
0$. For the derivations that follow, it is important to recall that, in KDE,
the kernel $K_i$ has integral equal to one. For the samples
$\{\xi^{(m)}\}_{m=1}^M \subset \R^d$ the KDE at a point $\xi =
(\xi_1,\xi_2,\dots,\xi_d)$ is given by
\begin{align}\label{KDEpdf}
p(\xi) &= \frac{1}{M h^d} \sum_{m=1}^M \mathcal{K}\left( \frac{\xi - \xi^{(m)}}{h} \right) \\ \nonumber
&= \frac{1}{M h^d} \sum_{m=1}^M K_1\left( \frac{\xi_1 - \xi^{(m)}_1}{h} \right) K_2\left( \frac{\xi_2 - \xi^{(m)}_2}{h}\right)
 \cdots K_d\left( \frac{\xi_d - \xi^{(m)}_d}{h} \right). \nonumber
\end{align}

The goal is to now use Equation (\ref{KDEpdf}) to build the conditional
cumulative distributions in the functions (\ref{rosenblatt}). The cumulative
distribution functions are built from the marginal densities,
\begin{align}\label{KDEmarginal}
p_{1,\dots,n}(\xi_1, \dots, \xi_n) &= \int p(\xi) \,d\xi_{n+1} \cdots d\xi_d \\ \nonumber
&= \frac{1}{M h^n} \sum_{m=1}^M K_1\left( \frac{\xi_1 - \xi^{(m)}_1}{h} \right) \cdots K_n\left( \frac{\xi_n -
    \xi^{(m)}_n}{h} \right). \nonumber
\end{align}
This leads directly to a formula for the marginal conditional cumulative
distributions
\begin{align}
F_{n|n-1,\dots,1}(\xi_n | \xi_1,\dots,\xi_{n-1}) &= \int_{-\infty}^{\xi_n} p_{n|n-1,\dots,1}(\tilde{\xi}_n |
\xi_1,\dots,\xi_{n-1}) \,d\tilde{\xi}_n \\ \nonumber
&= \int_{-\infty}^{\xi_n} \frac{p_{1,\dots,n}(\xi_1,\dots,\tilde{\xi}_n)}{p_{1,\dots,n-1}(\xi_1,\dots,\xi_{n-1})} \,
d\tilde{\xi}_n \\ \nonumber
&= \frac{\sum_{m=1}^M \left[ \left\{ \prod_{l=1}^{n-1} K_l \left( \frac{\xi_l - \xi_l^{(m)}}{h} \right)\right\} \frac{1}{h}
    \int_{-\infty}^{\xi_n} K_n \left( \frac{\tilde{\xi}_n - \xi_n^{(m)}}{h} \right) \, d\tilde{\xi}_n
  \right]}{\sum_{m=1}^M \left\{ \prod_{l=1}^{n-1} K_l \left( \frac{\xi_l - \xi_l^{(m)}}{h} \right) \right\}}. \nonumber
\end{align}

Once these can be computed, the inverse Rosenblatt transform is easily
accomplished. For each sample ${\bf u} \sim U[0,1]^d$, one iteratively goes
through the dimensions starting by computing $F_1(\xi_1)$ for increasing
$\xi_1$ until $F_1(\xi_1) \ge u_1$, which fixes a $\xi_1$. Next compute
$F_{2|1}(\xi_2 | \xi_1)$ for increasing $\xi_2$ until $F_{2|1}(\xi_2 | \xi_1)
\ge u_2$, which fixes $\xi_2$. This process is repeated $d$ times until we
arrive at a map $g({\bf u}) = \xi$.

Now that we have a method of computing the expectations in the polynomial
chaos coefficient definitions (\ref{PCcoefproj}), it is possible for us to
build an emulator from our data that combine the Karhunen-Loeve decomposition
in Equation (\ref{KLdecomposition}) and the polynomial chaos expansion in
Equation (\ref{PCexpansion}). This gives us an approximate model for the
stochastic process represented in the simulation,
\begin{equation}\label{KLPCdecomp}
  X(\theta; \omega(\zeta)) \approx \overline{X} + \sum_{n=1}^N \left( \sum_{k=1}^K c_{nk}(\theta)
    \Psi_k(\zeta) \right) f^n.
\end{equation}
The PC expansion must be computed separately for each value of our input
parameter set, $\theta$. The dependence of the stochastic process on the input
variables is then only seen through the coefficients $c_{nk}(\theta)$ in the
PC expansion. One important property of this emulation is to provide a map
from the multivariate standard normal random variable $\zeta$ to the intrinsic
uncertainty in the random vector $\xi(\theta; \omega)$. This defines a
probability space for the intrinsic uncertainty in the simulation that can be
sampled quickly. Once the emulator is constructed, for a fixed set of inputs,
many realizations may be computed to ascertain the approximate distributions
associated with quantities of interest when only intrinsic uncertainty is
present. The use of the polynomial chaos decomposition allows this intrinsic
uncertainty to be represented in a non-Gaussian, and somewhat, non-parametric
way which has the potential to represent very general behavior in the
intrinsic variability.

\begin{figure}[h]
  \begin{center}
    \includegraphics[scale=.4]{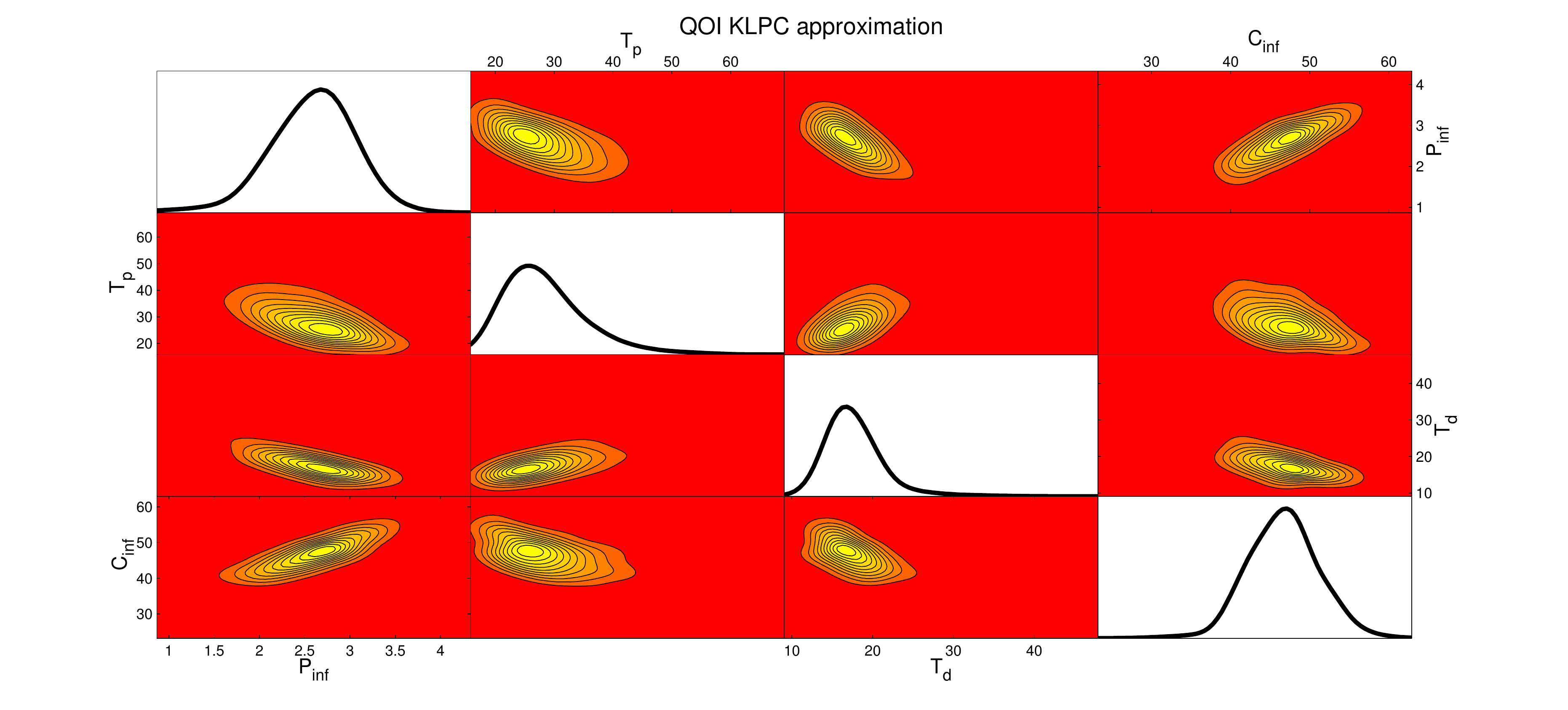}
  \end{center}
  \caption{Reconstruction of the one and two dimensional marginal
    distributions of the QOI using the KL-PC decomposition in
    (\ref{KLPCdecomp}). This figure should be compared with the original QOI
    distribution depicted in Figure \ref{fig:QOI_trueKDE}. Notice that much of
    the original distribution structure is maintained. However, the
    approximation inevitably smooths the original distribution and cuts off
    some of the lower probability regions. This is due to a mixture of the
    effects of truncation in the KL-PC decomposition and the bandwidth chosen
    in the KDE used to compute the PC coefficients.}
  \label{fig:KLPC_QOIrecon}
\end{figure}

An approximation of Equation (\ref{KLPCdecomp}) is depicted, for the SIR
example, in Figure \ref{fig:KLPC_QOIrecon}. Again, we see the smoothing out of
the QOI distribution's finer scale features. However, general shape,
correlation structure, and non-Gaussianity are preserved well. This
approximation can be made exact by increasing the number of samples from the
SIR model used to form the KL decomposition and the PC decomposition as well
as including more terms in each of the decompositions.

\section{Gaussian process regression on PC coefficients}

It remains to provide a surrogate model for the dependence of the stochastic
process on the parameterization, $\theta$. From the KL-PC approximation in
(\ref{KLPCdecomp}) for each set of inputs, we arrive at a set of $N \!\! \cdot
\! K$ coefficients $c_{nk}(\theta)$. We perform Gaussian process (GP)
regression on the coefficients with the samples from sets of input parameters
as data. There are many good reviews on the advantages of Gaussian processes
in statistical modeling \cite{mackay1998introduction,neal1997monte}. We
highlight two properties that are particularly important. First, a Gaussian
process regression is an interpolant. If the coefficients are computed at a
specific $\theta$ value and a Gaussian process is formed for the coefficients,
then the GP coefficient values will equal the observed coefficients at that
$\theta$ value. Second, a Gaussian process regression fits a process to data
through a series of observed input values, which is done in a way that ensures
the variance in the process will grow for $\theta$ values that are farther
away from observations. This allows uncertainty in the emulator to emerge that
is attributable to lack of observations, realizations taken at too few sets of
input parameters, and to describe the actual dependencies of the simulation on
input parameters.

To build the GP regression, we follow the methods introduced in
\cite{higdon2008computer,williams2006combining}, the necessary details are included for
completeness. First, we form a vector of the coefficients,
\begin{equation}
\vec{c}(\theta) = (c_{11}(\theta), c_{12}(\theta), \dots, c_{21}(\theta), c_{22}(\theta), \dots, c_{N(K-1)}(\theta), 
                   c_{NK}(\theta))^T.
\end{equation}
The statistical surrogate model is then built for $\vec{c} : \R^p \rightarrow
\R^{N \cdot K}$. If the PC coefficients are constructed (i.e., sampled) at $m$
parametric values $\{ \theta_1, \theta_2, \dots, \theta_m \}$, then we can
form the $(NK) \times m$ data matrix,
\begin{equation}
\Ct = \left( \begin{array}{cccc}
c_{11}(\theta_1) & c_{11}(\theta_2) & \dots & c_{11}(\theta_m) \\
c_{12}(\theta_1) & c_{12}(\theta_2) & \dots & c_{12}(\theta_m) \\
\vdots & \vdots & \ddots & \vdots \\
c_{NK}(\theta_1) & c_{NK}(\theta_2) & \dots & c_{NK}(\theta_m)
\end{array} \right).
\end{equation}
This is then standardized by subtracting the mean over all the samples and
dividing by the standard deviation of all the coefficients to form
$\Ct_{\textrm{std}}$. This standardized data matrix is then decomposed using a
singular value decomposition. In addition, a truncated $(NK) \times p_c$
matrix of singular vectors is constructed, $K = [{\bf k}_1, {\bf k}_2, \dots,
{\bf k}_{p_c}]$ with ${\bf k}_i \in \R^{N \cdot K}$. We truncate the singular
vectors to keep only those corresponding to large singular
values. $\Ct_{\textrm{std}}$ is used to build an emulator, corresponding to a
standardized map $\vec{c}_{\textrm{std}} : \R^p \rightarrow \R^{N \cdot K}$,
of the form
\begin{align}\label{std_gpmodel}
\vec{c}_{\textrm{std}}(\theta) &\approx \sum_{i=1}^{p_c} {\bf k}_i w_i(\theta; \eta) + \delta(\eta) \\ \nonumber
&= K W(\theta; \eta) + \delta(\eta)
\end{align}
where $W(\theta; \eta) = (w_1(\theta; \eta), \dots, w_{p_c}(\theta; \eta))^T$
and $\delta(\eta)$ is an independent zero mean Gaussian process modeling the
discrepancy between the truncated decomposition and the data. This is taken to
be $\delta(\eta) \sim \Norm ({\bf 0}_{NK}, \lambda_{\delta}^{-1} I_{NK})$ with
hyperparameter $\lambda_{\delta}$. The parameter $\lambda_{\delta}$ will
control how much noise is present, we will refer to $\lambda_{\delta}$ as the
\emph{noise precision} parameter. Here $w_i(\theta; \eta)$ will be $p_c$
independent zero mean Gaussian processes over the input space $\theta \in
\R^p$. These will be constructed from the $p_c \times m$ matrix
\begin{equation}
\mathcal{W} = \left( \begin{array}{cccc}
w_1(\theta_1) & w_1(\theta_2) & \dots & w_1(\theta_m) \\
w_2(\theta_1) & w_2(\theta_2) & \dots & w_2(\theta_m) \\
\vdots & \vdots & \ddots & \vdots \\
w_{p_c}(\theta_1) & w_{p_c}(\theta_2) & \dots & w_{p_c}(\theta_m)
\end{array} \right)
\end{equation}
coming from the truncated singular value decomposition, $\Ct_{\textrm{std}} =
K \mathcal{W}$.

Each of the $w_i(\theta; \eta)$ have covariance model
\cite{higdon2008computer}
\begin{equation}\label{cov_model}
R(\theta, \theta') = \frac{1}{\lambda_{w_i}} \prod_{k=1}^p \rho_{w_i(k)}^{4 (\theta_{(k)} - \theta'_{(k)})^2}
\end{equation}
with hyperparameters $\lambda_{w_i}$ and $\vec{\rho}_{w_i} = (\rho_{w_i(1)},
\dots, \rho_{w_i(p)})^T$. We will choose prior distributions on the
hyperparameters that ensure $\rho_{w_i(k)} \in (0,1)$ so that $R(\theta,
\theta')$ decays as $\| \theta - \theta' \| \rightarrow \infty$. The notation
$\theta_{(k)}$ is used to denote the $k^{th}$ coordinate of $\theta \in
\R^p$. For the Bayesian regression, we define the length $m$ vector ${\bf w}_i
= (w_i(\theta_1), w_i(\theta_2), \dots, w_i(\theta_m))^T$ for $i = 1, 2,
\dots, p_c$. The vector ${\bf w}_i$ has covariance given by
\begin{equation}
\lambda_{w_i}^{-1} C(\theta; \vec{\rho}_{w_i}).
\end{equation}

A symmetric $m \times m$ matrix is obtained by applying the covariance model
(\ref{cov_model}) to each pair in the sample set $\{ \theta_1, \dots, \theta_m
\}$.

Now we define the $m p_c$ vector of all processes ${\bf w}_i$ evaluated at the
sample points, $\vec{{\bf w}} = ({\bf w}_1^T, {\bf w}_2^T, \dots, {\bf
  w}_{p_c}^T)^T$, distributed as
\begin{align}\label{w_dist}
\vec{{\bf w}} &\sim \Norm({\bf 0}_{m p_c}, \Sigma_{{\bf w}} + \lambda_{\delta}^{-1} I_{m p_c}) \\
\Sigma_{{\bf w}} &= \diag_{i = 1,\dots,p_c}(\lambda_{w_i}^{-1} C(\theta; \vec{\rho}_{w_i})), 
\end{align}
the covariance matrix being size $(m p_c) \times (m p_c)$. 

Then the $m \cdot (N K)$ column vector of the standardized data
\begin{equation*}
  \vec{{\bf c}}_{\textrm{std}} = (\vec{c}_{\textrm{std}}(\theta_1)^T,
  \vec{c}_{\textrm{std}}(\theta_2)^T, \dots, \vec{c}_{\textrm{std}}(\theta_m)^T)^T
\end{equation*}
is distributed
\begin{equation}\label{c_dist}
\vec{{\bf c}}_{\textrm{std}} \sim \Norm( {\bf 0}_{m \cdot (N K)}, \tilde{K} \Sigma_{{\bf w}} \tilde{K}^T  +
\lambda_{\delta}^{-1} I_{m \cdot (N K)}).
\end{equation}
Where $\tilde{K} = [ I_m \otimes {\bf k}_1, I_m \otimes {\bf k}_2, \dots, I_m
\otimes {\bf k}_{p_c}]$ is the $(m \cdot (N K)) \times (m p_c)$ matrix formed
by the Kronecker product of the principle vectors. Notice that from Equation
(\ref{std_gpmodel}), it follows that $\tilde{K}^T \vec{{\bf c}}_{\textrm{std}}
= \vec{{\bf w}}$.

Relations (\ref{std_gpmodel}) and (\ref{w_dist})-(\ref{c_dist}) define a
likelihood coming from the density of the normal distribution
\begin{equation}
L(\vec{{\bf w}} | \lambda_{\delta}, \lambda_{w_i}, \vec{\rho}_{w_i}, i = 1, \dots, p_c). 
\end{equation}
In the regression step the posterior distribution is given by
\begin{align}\label{posterior}
p(\lambda_{\delta}, \lambda_{w_i}, \vec{\rho}_{w_i}, i = 1, \dots, p_c | \vec{{\bf w}}) &\propto \\ \nonumber
& L(\vec{{\bf w}} | \lambda_{\delta}, \lambda_{w_i}, \vec{\rho}_{w_i}, i = 1, \dots, p_c)
\pi(\lambda_{\delta}) \prod_{i=1}^{p_c} \left\{ \pi(\lambda_{w_i}) \prod_{k=1}^p \pi(\rho_{w_i(k)}) \right\}.
\end{align}

This is sampled using a Metropolis-Hastings MCMC \cite{robert1999monte} method
such as a univariate random walk or an independence sampler. One can either
take the estimated maximum likelihood values for the hyperparameters
$\lambda_{\delta}$, $\lambda_{w_i}$, $\vec{\rho}_{w_i}$ or choose a range of
values from the samples.

We assume the $\lambda$ hyperparameters have prior gamma distributions and the
$\rho$ hyperparameters have beta distribution priors
\cite{higdon2008computer},
\begin{align}\label{gp_priors}
\pi(\lambda_{w_i}) &\propto \lambda_{w_i}^{a_w-1} e^{-b_w \lambda_{w_i}}, \, i = 1, 2, \dots, p_c \\ \nonumber
\pi(\rho_{w_i(k)}) &\propto \rho_{w_i(k)}^{a_{\rho} - 1} (1 - \rho_{w_i(k)})^{b_{\rho} - 1}, \, i = 1, 2, \dots, p_c, \,
k = 1, 2, \dots, p \\ \nonumber
\pi(\lambda_{\delta}) &\propto \lambda_{\delta}^{a_{\delta}-1} e^{-b_{\delta} \lambda_{\delta}}.
\end{align}

Once values of the hyperparameters are chosen through exploration of the
posterior (\ref{posterior}), predictions can be made for the emulated
$\vec{c}_{\textrm{std}}(\theta^*)$ at some new parameter set $\theta^* \in
\R^p$. Let the prediction vector, at a set of parameter values $\{ \theta^*_1,
\theta^*_2, \dots, \theta^*_s \}$, be denoted $\vec{{\bf c}}^*_{\textrm{std}}
= (\vec{c}_{\textrm{std}}(\theta^*_1)^T, \vec{c}_{\textrm{std}}(\theta^*_2)^T,
\dots, \vec{c}_{\textrm{std}}(\theta^*_s)^T)^T$, a length $s \cdot (N K)$
column vector. The prediction is based on prediction of the $s \cdot p_c$
vector $\vec{{\bf w}}^* = (({\bf w}^*_1)^T, ({\bf w}^*_2)^T, \dots, ({\bf
  w}^*_{p_c})^T)^T$, where ${\bf w}^*_i = (w_i(\theta^*_1), w_i(\theta^*_2),
\dots, w_i(\theta^*_s))^T$ for $i = 1, 2, \dots, p_c$.

From our definitions above we see that the data vector and prediction vector
are jointly distributed as
\begin{equation}
\left[
\begin{array}{c}
\vec{{\bf w}} \\
\vec{{\bf w}}^*
\end{array} \right]
\sim \Norm (0_{(s+m) p_c}, \Xi + \lambda_{\delta}^{-1} I_{(s+m) p_c}), \, \Xi = \left[
\begin{array}{cc}
\Sigma_{{\bf w}} & \Sigma_{{\bf ww^*}} \\
\Sigma_{{\bf w^*w}} & \Sigma_{{\bf w^*}}
\end{array} \right].
\end{equation}
The terms in the covariance matrix come from applying our covariance model
(\ref{cov_model}) to each pair of the respective parameter sets. Thus,
$\Sigma_{{\bf w}}$ is the $m \cdot p_c$ square matrix obtained by applying
Equation (\ref{cov_model}) to each pair in the \emph{sample} set $\{ \theta_1,
\dots, \theta_m\}$ for each ${\bf w}_i$. Similarly, $\Sigma_{{\bf w^*}}$ is
the $s \cdot p_c$ square matrix obtained by applying Equation
(\ref{cov_model}) to each pair in the \emph{prediction} set $\{ \theta^*_1,
\dots, \theta^*_s\}$ for each ${\bf w}^*_i$. Finally, $\Sigma^T_{{\bf ww^*}} =
\Sigma_{{\bf w^*w}}$ is the $(s \cdot p_c) \times (m \cdot p_c)$ matrix
obtained by applying Equation (\ref{cov_model}) to each pair $(\theta_k,
\theta^*_j)$ in the sample and prediction set. The predictions are then
distributed as follows:
\begin{align}
\vec{{\bf w}}^* &\sim \Norm ( \mu^* , \Omega^*) \\ \nonumber
\mu^* &= \Sigma_{{\bf w^*w}} \left( \Sigma_{{\bf w}} + \lambda_{\delta}^{-1} I_{m p_c} \right)^{-1} \vec{{\bf w}} \\ \nonumber
\Omega^* &= \left( \Sigma_{{\bf w^*}} + \lambda_{\delta}^{-1} I_{s p_c} \right) - \Sigma_{{\bf w^*w}} \left(
  \Sigma_{{\bf w}} + \lambda_{\delta}^{-1} I_{m p_c} \right)^{-1} \Sigma_{{\bf ww^*}}.
\end{align}

From the predictions $\vec{{\bf w}}^*$, we can define predicted standardized
polynomial chaos coefficients using the relations
\begin{align}
\vec{{\bf c}}^*_{\textrm{std}} &= \tilde{K}^* \vec{{\bf w}}^* \\
\tilde{K}^* &= [ I_s \otimes {\bf k}_1, I_s \otimes {\bf k}_2, \dots, I_s \otimes {\bf k}_{p_c}].
\end{align}
Thus, after destandardization, we have defined a new map through Bayesian
Gaussian process regression for the coefficients denoted by
\begin{equation}
\tilde{c}(\theta; \eta) = (\tilde{c}_{11}(\theta; \eta), \tilde{c}_{12}(\theta; \eta), \dots, \tilde{c}_{21}(\theta;
\eta), \dots , \tilde{c}_{NK}(\theta; \eta))^T.
\end{equation}
The random variable $\eta$ is the associated state space variable for the
Gaussian process. For a given realization of the GP regression $\eta$ is
fixed.

Once the GP is formed for the PC coefficients, we build a complete emulator of
the intrinsic and parametric uncertainty in the simulation. The final form of
the statistical surrogate is
\begin{equation}\label{KLPCGPemulator}
  \mathcal{X}^e(\theta; \zeta, \eta) = \overline{X} + \sum_{n=1}^N \left( \sum_{k=1}^K \tilde{c}_{nk}(\theta,\eta)
    \Psi_k(\zeta) \right) f^n.
\end{equation}

\begin{figure}[h]
  \begin{center}
      \includegraphics[scale=.4]{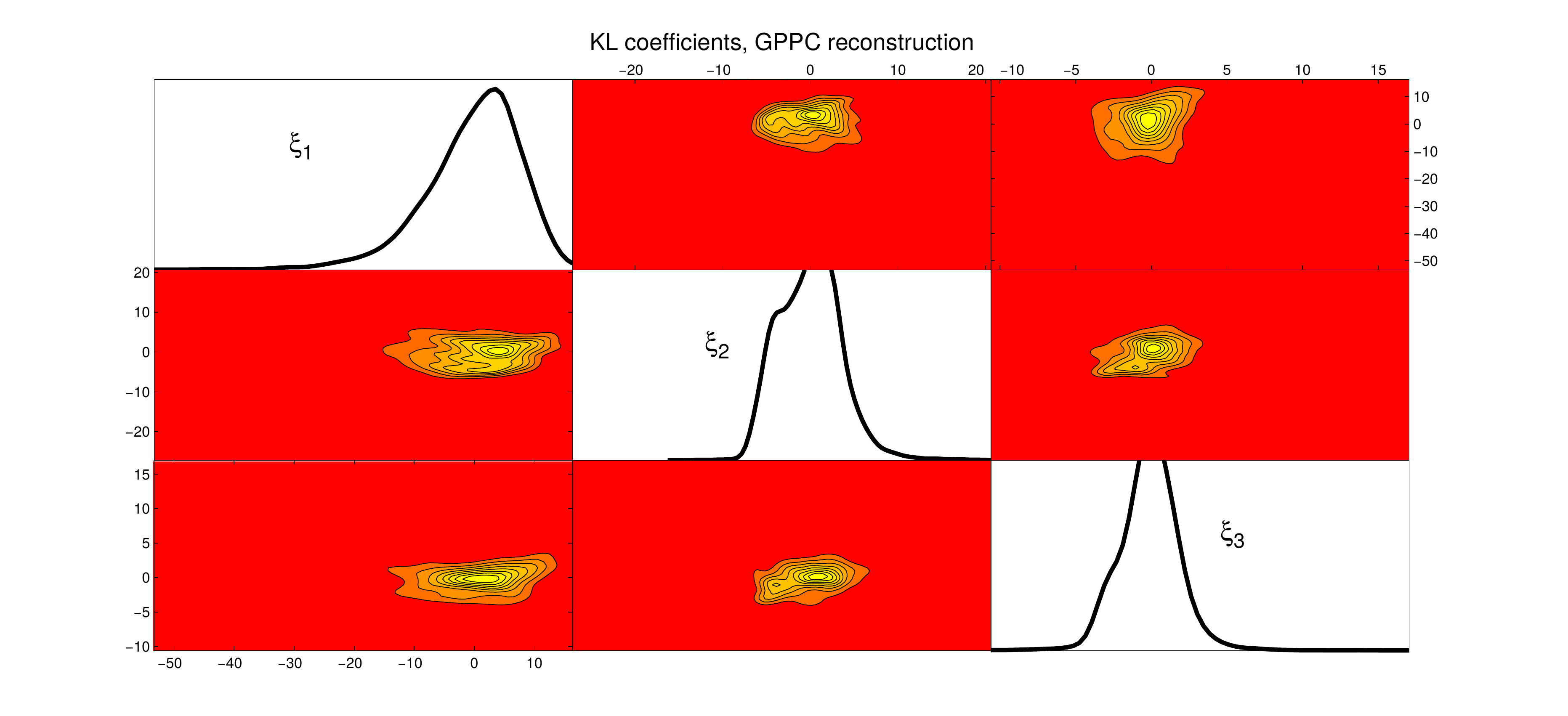}
  \end{center}
  \caption{One and two dimensional marginal distributions of the KL
    coefficients for the SIR QOI. These were reconstructed using Gaussian
    process regression on the polynomial chaos coefficients depicted in Figure
    \ref{fig:KLcoefPCrecon} using equation (\ref{KLPCGPemulator}).}
  \label{fig:QOI_GPPCrecon}
\end{figure} 

\begin{figure}[h]
  \begin{center}
      \includegraphics[scale=.4]{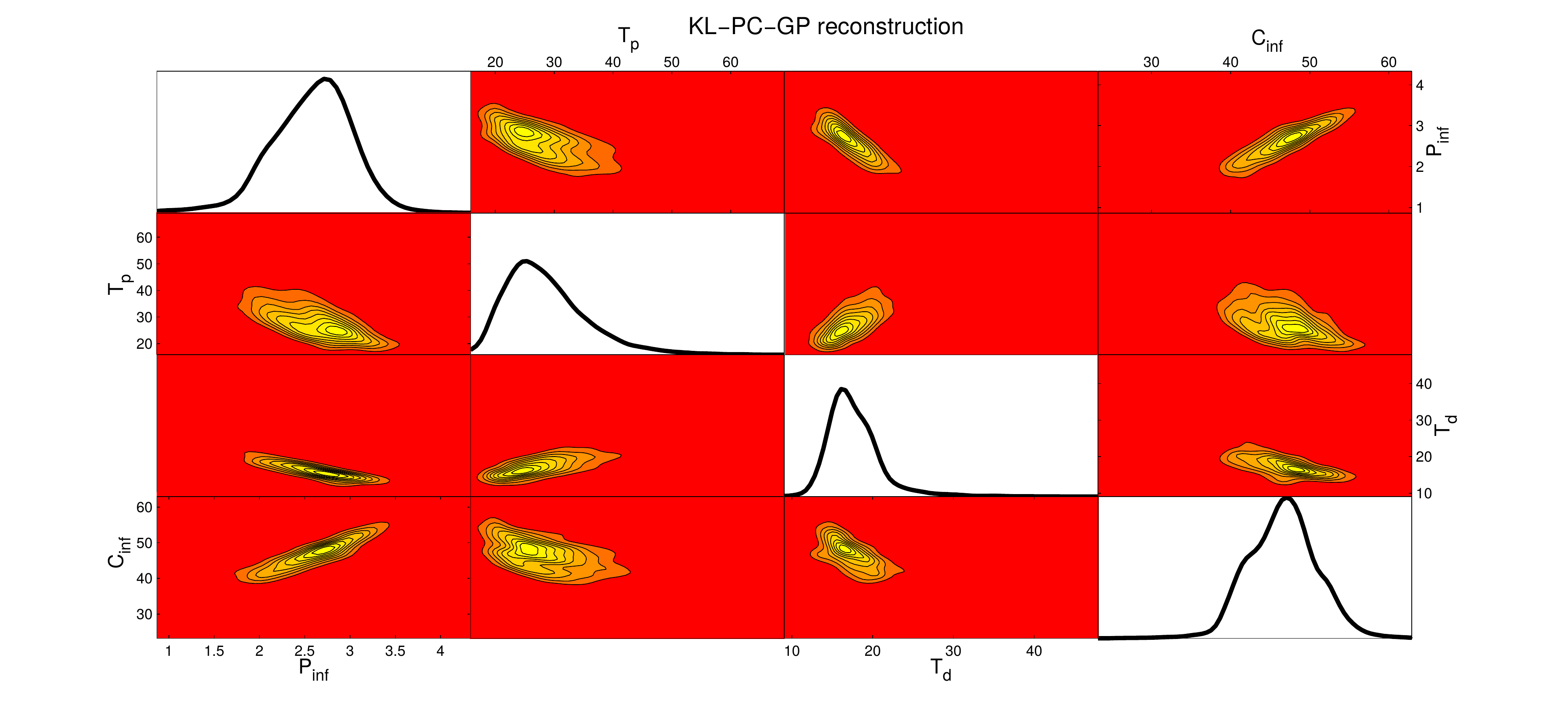}
  \end{center}
  \caption{One and two dimensional marginal distributions of the SIR QOI
    reconstructed using our combination of Karhunen-Loeve, polynomial chaos,
    and Gaussian process regression. This is an approximation of the original
    distribution in Figure \ref{fig:QOI_trueKDE}. The approximation captures
    the overall shape and range of the original QOI. However, the
    approximation does concentrate more around the mode of the original
    distribution. As sample size increases this effect will diminish.}
  \label{fig:QOI_KLPCGPrecon}
\end{figure} 

In Equation (\ref{KLPCGPemulator}), the contributions from each of the
separate sources of uncertainty is represented explicitly. The first source of
variation is the change between QOI. This variation is controlled in the
emulator through the KL decomposition parameter $\tau$. Intrinsic variation is
represented in $\mathcal{X}^e$ by the standard multivariate normal random
variable $\zeta$ arising from the PC expansion. The parameter $\theta$ occurs
in the GP of the coefficients and controls the dependence of the emulator on
the input parameters. Lastly, the use of a GP regression to relate PC
coefficients at distinct $\theta$ values introduces a new source of
uncertainty that is not originally present in the model. This is uncertainty
in the emulator's dependence on the inputs that has arisen from lack of
sampling in $\theta$. The random variable $\eta$ now quantifies this
uncertainty, giving a way to sample from the emulator that will inform how
effective the input evaluation scheme was.

In using $\mathcal{X}^e$ to compute approximate probabilities of specific
outcomes from the simulation, each of these variables can be sampled
independently. This permits independent study of each source of variation in
the model to quantify its effects on the simulation predictions. Moreover, it
allows one to study regions of the input space where surrogate model
discrepancy is largest, i.e., where $\eta$ has the biggest contribution. This
provides a framework to determine what simulation realizations would improve
$\mathcal{X}^e$ the most.

We use samples of $\vec{Q}$ to compute a statistical emulator for the
stochastic SIR model. We denote the emulation by
\begin{align}
\vec{Q}_e (\beta, \gamma; \eta, \zeta) &= \E[\vec{Q}] + \sum_{n=1}^4 \left( \sum_{k=1}^K \tilde{c}_{nk}(\beta, \gamma; \eta)
    \Psi_k(\zeta) \right) \vec{f}_n \\
\beta \sim \ln \Norm(\mu_{\beta}, \sigma^2_{\beta}), &\,\, \gamma \sim \ln \Norm(\mu_{\gamma}, \sigma^2_{\gamma}),\,\, 
\zeta \sim \Norm(0_4, I_4).
\end{align}
The reconstruction of the first three KL coefficients using Gaussian process
emulation of the PC coefficients in our KL decomposition is shown in Figure
\ref{fig:QOI_GPPCrecon}. Then the KL reconstruction can be used to approximate
the original QOI random variable. The distribution of this approximating
random variable is depicted in Figure \ref{fig:QOI_KLPCGPrecon}. Overall the
shape of the distribution is maintained but more importantly, by using this
approximation, we have the ability to sample from each source of uncertainty
separately.

Recall, $\eta$ represents the state variable of the Gaussian process
$\tilde{c}_{nk}$. The variables $(\beta, \gamma)$ can be sampled to explore
the uncertainty in $\vec{Q}_e$ due to uncertainty in the input
parameters. Similarly, $\zeta$ can be sampled to explore the uncertainty
contribution due to intrinsic variation in the model. Finally, we can take
many realizations of the Gaussian process $\tilde{c}_{nk}$ to study the
uncertainty in the emulator $\vec{Q}_e$ introduced by lack of sampling of the
actual simulation $\vec{Q}$.

Since realizations of the emulator $\vec{Q}_e$ are fast, we can use a Monte
Carlo method combined with KDE to reconstruct an approximation of the
distribution of $\vec{Q}$. This distribution can then be used to estimate
probabilities of events associated with the quantities of interest in
$\vec{Q}$.

\section{Conclusion}
 
We have discussed the problem of propagating uncertainty through a simulation
that yields predictions effected by both intrinsic and parametric
uncertainty. The term parametric uncertainty was used to characterize sources
of variation in the simulation's predictions for which the researcher knows
the associated probability space and has control of how it is sampled when
running the simulation. Intrinsic uncertainty was defined as variation in the
simulation's predictions that lacked an underlying parameterized probability
space and/or a source of variation the researcher did not have control over.

The statistical emulator constructed yielded a parametrization of the
intrinsic probability space that allowed for non-Gaussianity and separated the
different sources of uncertainty in the output. This approach included
accounting for uncertainty introduced from lack of sampling the simulation.

We used a stochastic system of ordinary differential equations, representing a
simple disease model, to illustrate our methods.  We allowed the input
parameters of infection rate and recovery rate to be random variables with
lognormal distributions. The four quantities of interest, the peak infected
percent of the population, the time to the peak of the epidemic, the duration
of the outbreak, and the total percent of the population infected, were then
emulated from samples.

The emulation methods have potential application to uncertainty quantification
in large scale computer simulations. Their utility can be evaluated by their
ability to reconstruct joint distributions of QOI accurately from samples of
the simulation. However, the emulation method has a large variety of tuning
parameters and therefore requires human supervision in its application.  It
would, therefore, be of great utility to have reliable heuristics and theorems
that give precise hypotheses on the types of stochastic processes for which
the emulator will converge and the emulator's rate of convergence, based on
sampling size of the simulation. This is especially true in the case of
stochastic processes whose distribution at a given parameter set depend on the
parameter values. In the absence of theorems on convergence and rates of
convergence for statistical surrogates, one would like to have a good suite of
numerical tools at their disposal to evaluate the performance of the
emulator. Such methods would need to be non-parametric and non-intrusive to be
able to be applied to a large range of simulations without \emph{a priori}
assumptions. In the future, we will pursue these types of convergence results
and numerical evaluation methods.

\section*{Acknowledgements} This research has been supported at Los Alamos
National Laboratory under the Department of Energy contract DE-AC52-06NA25396
and a grant from the NIH/NIGMS in the Models of Infectious Disease Agent Study
(MIDAS) program U01-GM097658-01. The authors thank Dave Higdon, Nick
Hengartner, Reid Priedhorsky, Geoff Fairchild, Nicholas Generous, and Susan
Mniszewski at Los Alamos National Laboratory and Carrie Manore at Tulane
University for providing valuable comments on this work. This work has been
approved for release under LA-UR-15-20044.

\bibliographystyle{plain}
\bibliography{SIR_UQ}

\end{document}